\documentclass[journal]{vgtc}                     

\onlineid{1480}

\ieeedoi{10.1109/TVCG.2023.3247054}

\vgtccategory{Research}
\vgtcpapertype{algorithm/technique}

\title{PACE: Data-Driven Virtual Agent Interaction in Dense and Cluttered Environments}

\author{\authororcid{James F. Mullen Jr}{0000-0002-4117-1741} and \authororcid{Dinesh Manocha}{0000-0001-7047-9801}}
\affiliation{University of Maryland College Park}

\authorfooter{
  \item
  	James Mullen E-mail: mullenj@umd.edu
  \item
  	Dinesh Manocha E-mail: dmanocha@umd.edu
}

\shortauthortitle{Mullen \MakeLowercase{\textit{et al.}}: Virtual Agent Interaction}

\abstract{We present PACE, a novel method for modifying motion-captured virtual agents to interact with and move throughout dense, cluttered 3D scenes. Our approach changes a given motion sequence of a virtual agent as needed to adjust to the obstacles and objects in the environment. We first take the individual frames of the motion sequence most important for modeling interactions with the scene and pair them with the relevant scene geometry, obstacles, and semantics such that interactions in the agents motion match the affordances of the scene (e.g., standing on a floor or sitting in a chair). We then optimize the motion of the human by directly altering the high-DOF pose at each frame in the motion to better account for the unique geometric constraints of the scene. Our formulation uses novel loss functions that maintain a realistic flow and natural-looking motion. We compare our method with prior motion generating techniques and highlight the benefits of our method with a perceptual study and physical plausibility metrics. Human raters preferred our method over the prior approaches. Specifically, they preferred our method 57.1\% of the time versus the state-of-the-art method using existing motions, and 81.0\% of the time versus a state-of-the-art motion synthesis method. Additionally, our method performs significantly higher on established physical plausibility and interaction metrics. Specifically, we outperform competing methods by over 1.2\% in terms of the non-collision metric and by over 18\% in terms of the contact metric. We have integrated our interactive system with Microsoft HoloLens and demonstrate its benefits in real-world indoor scenes. Our project website is available at \url{https://gamma.umd.edu/pace/}
} 


\teaser{
  \centering
  \includegraphics[width=\linewidth]{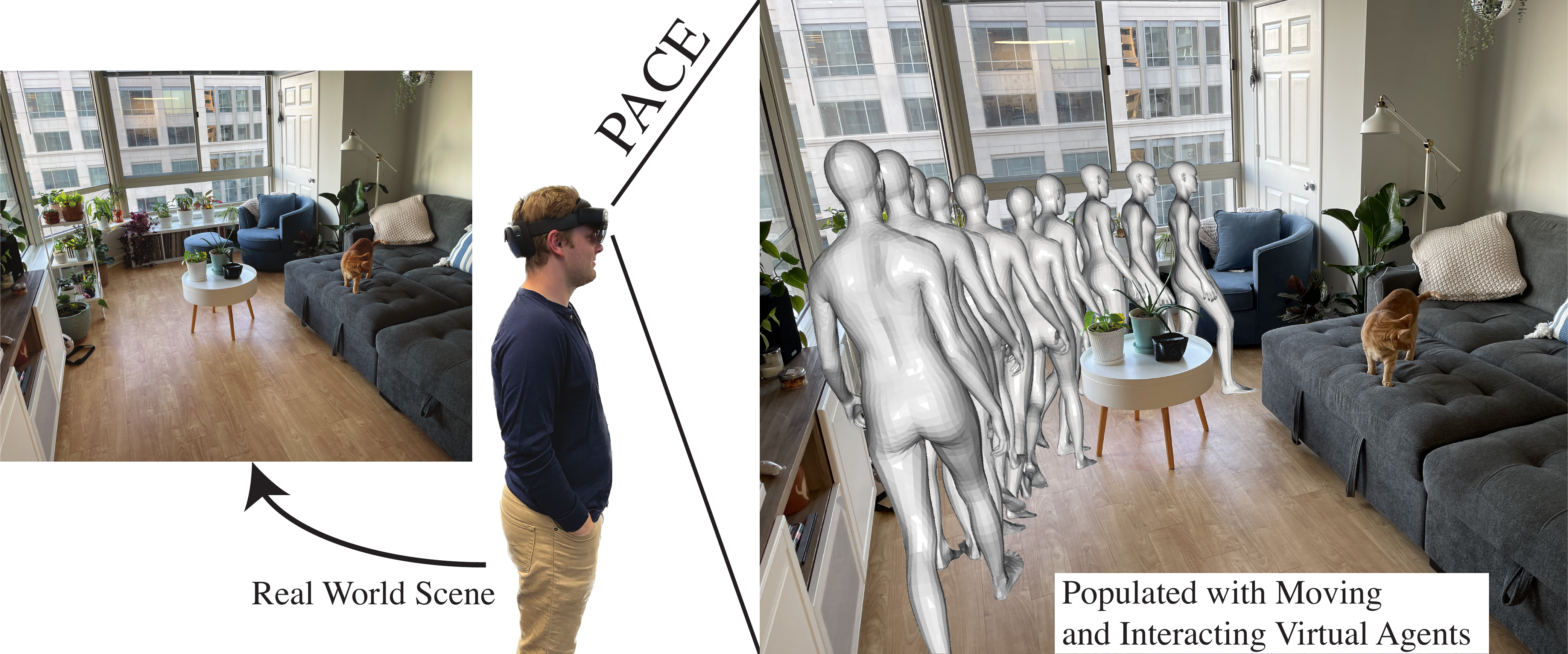}
  \caption{We present an approach to create moving virtual agents that interact with a dense or cluttered 3D scene, either virtual or real, based in motion-captured human data. We have integrated our system with Microsoft HoloLens and highlight its benefit in terms of moving virtual agents in real-world scenes. As key functionality of our method, we adjust or tailor the motion-captured agents motion to the obstacles in the scene like the chairs and tables in the scene above. Our optimization-based approach improves the performance in terms of well-known interaction metrics corresponding to non-collision and contact. In this figure, which we captured using our Microsoft HoloLens, we observe a virtual agent navigating around the obstacles in the environment.}
  \label{fig:front}
}

\graphicspath{{figs/}{figures/}{pictures/}{images/}{./}} 

\usepackage{tabu}                      
\usepackage{booktabs}                  
\usepackage{lipsum}                    
\usepackage{mwe}                       

\usepackage{mathptmx}                  

\usepackage{amsmath}

\begin{document}

\firstsection{Introduction}
\maketitle
Humans interact with their environment by making contact with objects and avoiding collisions with obstacles. For example, a human sitting at a desk is sitting in a chair. Their arms are likely resting on the surface of the desk, potentially even grasping the mouse and keyboard. When this human stands up from their desk, they have to navigate around the room, avoiding any objects within it, before reaching out to grasp the door knob and open the door. These types of interactions, defined by contact, dictate how humans move through their environment. In this paper, our goal is to take motion-captured humans and translate them into virtual humans that interact with virtual or real scenes in a natural looking way.

To motivate our problem, imagine a \textit{virtual} office. To make it feel plausible, virtual human agents or avatars have to be moving around the space and interacting with objects such as the chairs, desks, or keyboards with no penetrations or awkward motions. Current methods for this type of dense, cluttered space are based on animators using modeling or animation tools to generate such motion sequences that fit the environment. However, this can be very time consuming and it requires animators with considerable experience using these tools. Additionally, cost prohibits a large number of specialized animations or motions from being produced, resulting in the same animations being repeated in the virtual environment, reducing the sense of realism of the environment and the users engagement within it. In contrast, our goal is to leverage the vast quantities of publicly available human motion-captured data and place virtual agents personifying the motion-capture into the environment. 

Populating an environment with virtual humans that interact with the environment naturally is an important problem in virtual reality (VR), augmented reality (AR) \cite{hassanPopulating3DScenes2021}, game design \cite{martinHowMakeImmersive}, and human-robot interaction \cite{linVirtualRealityPlatform2016}. Specifically, a key problem in VR is generating plausible motion of human-like virtual agents. Prior methods are limited to humans moving in open or large spaces with few obstacles. However, real world environments, especially indoor ones, are filled with  obstacles and can be dense. Most prior work in VR in terms of dense or cluttered environments is mainly limited to object selection or manipulation \cite{narang2018simulating, shen2021simulating, bailenson2005independent}. Recent work in computer vision has enabled the creation of several real world datasets of indoor scenes containing dense objects and furniture\cite{Matterport3D, dai2017scannet, xiazamirhe2018gibsonenv, zillow, 3dfront, hassanResolving3DHuman2019}, such as Habitat-Matterport or Scan-Net, with hundreds of 3D scans of rooms with many objects. Such 3D models are increasingly used for VR and AR applications. However, we need to generate appropriate methods for virtual humans to interact with such environments by generating collision-free and plausible motion.  

Prior work in human-scene interaction, or scene affordances, attempts to place 3D human models into 3D scene scans such that the placement matches real human behavior. Most methods work with static, single-pose humans, \cite{kimShape2PoseHumancentricShape2014, hoSpatialRelationshipPreserving2010, zhangPLACEProximityLearning2020, hassanResolving3DHuman2019}. Recent methods enable the generalization of human model placement into any possible scene \cite{hassanPopulating3DScenes2021}, and extend this ability to existing short virtual agent motions \cite{mullenjrPlacingHumanAnimations2022}. Motion synthesis literature similarly tries to generate 3D human motion sequences with recent methods accounting for the scene geometry and affordances \cite{starkeNeuralStateMachine2019, lingCharacterControllersUsing2020, wangSynthesizingLongTerm3D2021}. 

\subsection{Main Contributions}
We leverage widely available indoor scene datasets and populate the complex, dense, indoor scenes with natural-looking virtual agents that move and interact with their environment. For the most natural-looking movements possible, we start with motion-captured human data and tailor it to the unique nature of each dense indoor scene. This is different from existing synthesis approaches that instead attempt to create motion sequences to fit the environment after learning from many valid sequences. When using motion-captured data, properly fitting a given environment as to make plausible interactions requires making changes to each pose in the motion sequence, each with several degrees of freedom, while maintaining the natural look and flow of the underlying motion sequence. We do this through a novel optimization method that takes the virtual agent motion sequence and 3D scene as inputs. From the scene and motion sequence, the optimization method outputs the location and orientation of the most natural placement of the virtual agent in the scene based on interaction metrics, and the modified pose and translation of the individual meshes in the virtual agent motion. While we did our evaluations with motion-captured data, we can utilize any virtual agent motion that can be represented as a time-series of 3D human meshes. We make no assumptions about the scene itself, but results are better if the objects in it are labeled. In practice, our approach shows the largest improvement over prior art with virtual agent motions longer than three seconds. Our main contributions are as follows:

\begin{enumerate}
\item We present PACE, a method to generate placements for motion-captured virtual agents into 3D dense scenes with motion optimization. Using existing methods, PACE first utilizes a deep model to estimate a virtual agent motion's potential scene interactions, before using this information alongside the motion geometry with another deep model to select important frames in the motion. Using the output of the deep models, we weight meshes in the motion such that the highest weight is attributed to meshes with maximum diversity and important scene interactions. This weighting allows us to find an optimal initial placement of the virtual agent before {our novel algorithms} optimize the agent's motion itself to better fit the unique constraints of a dense, cluttered scene. Our optimizer utilizes novel pose and motion losses alongside contact and semantic losses to alter the motion sequence to best fit the scene while maintaining the essence and natural look or the original motion.

\item We qualitatively show natural and physically plausible virtual agent placement. Through a perceptual user study we show that human raters prefer PACE 67.9\% of the time over an extension of POSA to the time dimension, 81.2\% of the time over a state-of-the-art motion synthesis method, and 59.3\% of the time over PAAK.

\item We quantitatively show a significant improvement over POSA\cite{hassanPopulating3DScenes2021} and PAAK\cite{mullenjrPlacingHumanAnimations2022} in the physical plausibility and interaction metrics\cite{zhangPLACEProximityLearning2020, hassanPopulating3DScenes2021}. These metrics include the non-collision and contact metrics which denote a lack of collisions and interactions with the scene respectively. Specifically, PACE improves upon PAAK by over 1.2\% in the non-collision metric and by over 18\% in the contact metric. 

\item We have integrated PACE with a Microsoft HoloLens and evaluate the performance in real world scenes. This allows users to populate their local environment with moving virtual humans that interact with the environment in natural-looking and physically plausible ways. The populated virtual agents move at 30 frames per second in a plausible manner.

\item We release a dataset with tens of thousands of virtual agents placed into dense and cluttered indoor scenes. We also release software to render videos from a camera perspective of the users choosing or interact with the entire 3D scene-agent pair.
\end{enumerate}
 
\section{Related Work}

\textbf{Generating Plausible Virtual Agent Motion.}
Closely related to our work is the problem of generating plausible virtual agent motion, especially in cluttered, dense virtual environments. This is a well studied problem \cite{ruddle2001movement, ruddle2004effect, lessels2005movement, simeone2017altering, argelaguet2009efficient}, with many trying to direct or maneuver \textit{a users} motion through a dense, cluttered environments. More recently, many researchers focused on plotting virtual agent motion through a virtual environment \cite{yePAVALPositionAwareVirtual2021, langVirtualAgentPositioning2019}, or generating virtual agents to fit an environment \cite{zhangGenerating3DPeople2020, liPuttingHumansScene2019}. We aim to combine many of these threads of research by creating a method capable of generating agent-scene pairings such that interactions with the environment are natural-looking and convincing for the user.

\textbf{Human Models.}
Most prior work with virtual human agents utilizes body skeletons to model 3D humans \cite{ionescuHuman36MLarge2014, sigalHumanEvaSynchronizedVideo2010}. However, the surface of the body is important for rending the virtual agent or modeling interactions with the scene or any objects. For example, it is essential to know how far the surface of the back is from the skeleton when placing an agent laying on a surface. Learned parametric 3D body models have addressed this need \cite{anguelovSCAPEShapeCompletion2005, jooTotalCapture3D2018, loperSMPLSkinnedMultiperson2015, pavlakosExpressiveBodyCapture2019, osmanSTARSparseTrained2020}. In our work, we utilize SMPL-X \cite{pavlakosExpressiveBodyCapture2019}, a body model that captures face and hand articulation in addition to the structure of the body itself.

\textbf{Motion Synthesis.} 
Motion synthesis is a well-studied problem in computer graphics, VR,  and computer vision \cite{kovarFlexibleAutomaticMotion2003, pavlovicLearningSwitchingLinear2000, urtasunTopologicallyconstrainedLatentVariable2008, harveyRobustMotionInbetweening2020, starkeNeuralStateMachine2019, lingCharacterControllersUsing2020, xuHierarchicalStylebasedNetworks2020, holdenDeepLearningFramework2016, rempeHuMoR3DHuman2021, wangSynthesizingLongTerm3D2021, wangDiverseNaturalSceneAware2022}. Early work in motion synthesis worked to synthesize intermediate states between two given points in the motion sequence \cite{urtasunTopologicallyconstrainedLatentVariable2008, harveyRobustMotionInbetweening2020, xiaLearningBasedSphereNonlinear2019}. These methods were known to not handle large translational position changes effectively. More recent work began utilizing data-driven deep models for motion synthesis \cite{xuHierarchicalStylebasedNetworks2020, holdenDeepLearningFramework2016}. These methods show better generalization than the geometric methods from earlier work. However, most of these methods look at the animation or generated motion in isolation from the scene or environment.

Some more recent motion synthesis methods account for both the environment and the virtual agent's movement \cite{starkeNeuralStateMachine2019, lingCharacterControllersUsing2020, cleggLearningDressSynthesizing2018, wangCriticRegularizedRegression2020}. Many of these methods use greatly simplified scenarios with predefined objects or primitive motion. Our work is closest to \cite{wangSynthesizingLongTerm3D2021} which utilizes arbitrary 3D environments when synthesizing its motion. The motions produced by \cite{wangSynthesizingLongTerm3D2021} and others frame our use case by falling short of the realism of motion-captured movements. However, our system, PACE does not synthesize its own motion, but instead leverages motion capture data and tailors it to the needs of the scene, resulting in the most natural-looking and physically plausible agent-scene pairing possible.

\begin{figure*}[ht]
	\begin{center}
		\includegraphics[width=\linewidth]{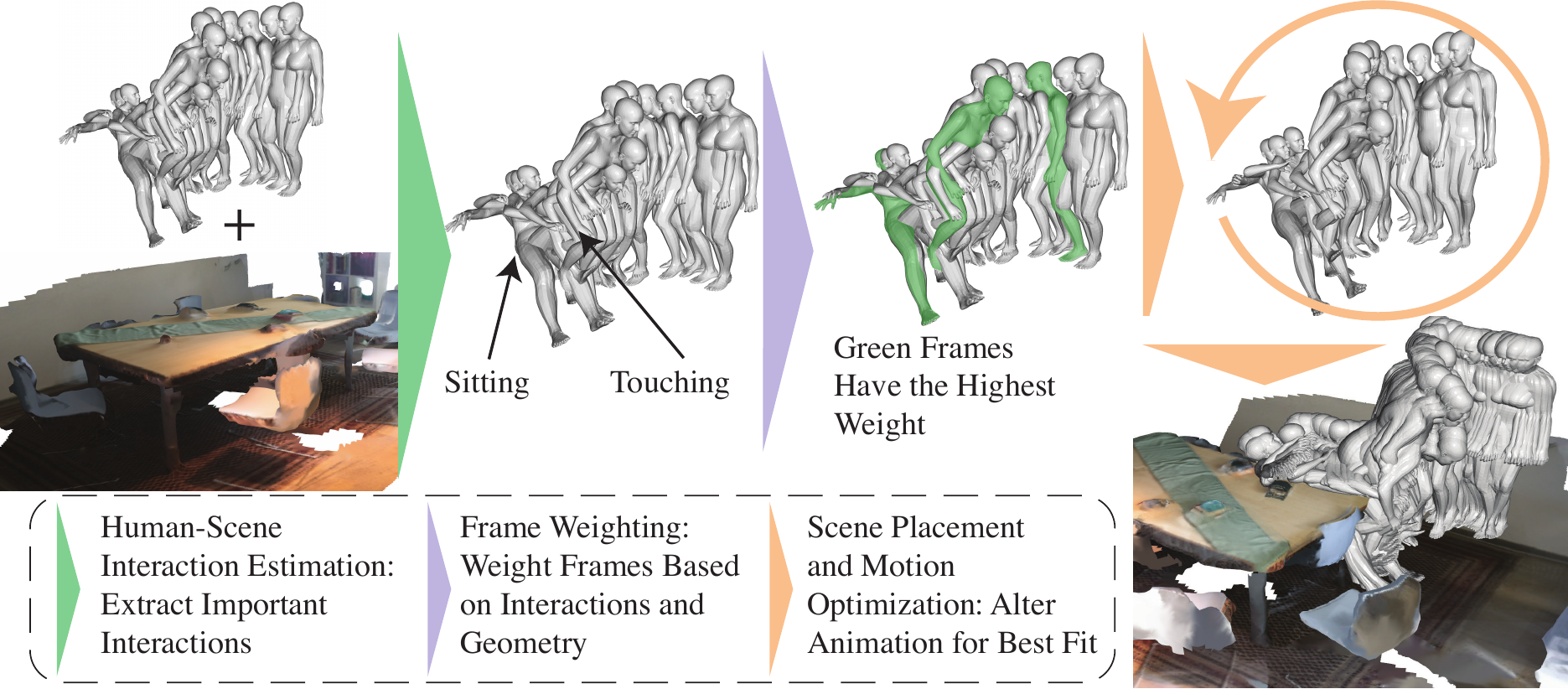}
        \caption{An overview of PACE. Using exiting methods, we first estimate human-scene interactions for a given scene and use those interactions to determine the frame weighting in the virtual agent motion. Through our novel contributions, we then utilize the scene geometry to optimize the motion of the virtual agent such that it interacts with the environment, matching both the interaction in the motion and the geometry of the scene based on our two interaction metrics: non-collision and contact.}
		\label{fig:method}
	\end{center}
\end{figure*}

\textbf{Video Synthesis.}
Our work is also related to the synthesis of videos containing human actions. Generative adversarial networks (GANs) \cite{goodfellowGenerativeAdversarialNets2014} and neural radiance fields (NeRFs) \cite{mildenhallNeRFRepresentingScenes2020} have contributed to a ever growing body of work attempting to generate videos of humans completing actions in arbitrary scenes. \cite{niemeyerGIRAFFERepresentingScenes2021, pengAnimatableNeuralRadiance2021, noguchiNeuralArticulatedRadiance2021, yangBANMoBuildingAnimatable2022, su2021anerf, yangViSERVideoSpecificSurface2021, guSTYLENERFSTYLEBASED3DAWARE2022, zhaoHumanNeRFEfficientlyGenerated2022, kwonNeuralHumanPerformer2021, Chan2021, noguchiUnsupervisedLearningEfficient2022} worked towards extending NeRF to scenes with multiple objects and arbitrary backgrounds, or to articulated objects like humans. Closest to our work is \cite{noguchiUnsupervisedLearningEfficient2022} which similarly takes a human action and attempts to place it on a background. Our approach addresses many of the shortcomings of \cite{noguchiUnsupervisedLearningEfficient2022}, enabling complex natural-looking background and natural-looking interactions between the animation and background.

\textbf{Human-Scene Interaction.}
Also closely related to our work is human-scene interaction (HSI), or scene affordance. Early efforts in HSI were purely geometric with Gleicher \cite{gleicherRetargettingMotionNew1998} and Kim et al. \cite{kimShape2PoseHumancentricShape2014} using contact constraints and automating 3D skeleton generation respectively. These and subsequent works exploited the importance of contact with the environment, with some accounting for the forces present in the environment \cite{kangEnvironmentAdaptiveContactPoses2014, leimerPoseSeatAutomated2020, grabnerWhatMakesChair2011}. As a major step forward, Gupta et al. \cite{gupta3DSceneGeometry2011} estimated the human poses "afforded" by the scene by predicting a 3D scene occupancy grid and computing the support of 3D skeletons inside of it.

\textbf{Data-Driven Methods.}
Many data-driven approaches have been proposed~\cite{tan2018, jiangHallucinatedHumansHidden2013, savvaSceneGrokInferringAction2014, zhangGenerating3DPeople2020, hassanPopulating3DScenes2021, liPuttingHumansScene2019}. \cite{jiangHallucinatedHumansHidden2013} and \cite{koppulaLearningHumanActivities2013} learn to estimate human poses and object affordances in an RGB-D 3D scene while \cite{wangGeometricPoseAffordance2021} learned to utilize the affordances of the scene to optimize pose estimation. Closest related to PACE, PSI \cite{zhangGenerating3DPeople2020}, PLACE \cite{zhangPLACEProximityLearning2020}, POSA \cite{hassanPopulating3DScenes2021}, and PAAK \cite{mullenjrPlacingHumanAnimations2022} populate scenes with SMPL-X \cite{pavlakosExpressiveBodyCapture2019} human bodies. Of these methods, POSA and PAAK are unique in that they are human-centric, finding estimated interactions in the human meshes and then pairing those interactions with any relevant affordances in the environment. In PACE, similar to PAAK, we leverage POSA's model and utilize the estimated contact and semantic information it provides about a given virtual agent motion sequence. Of all of these methods, PAAK is unique in its ability to place virtual agent motion instead of single, static human meshes. Our key difference from PAAK is our ability to tailor the virtual agent motions directly to the unique needs of a scene. PAAK is handcuffed by the geometry of the motion, resulting in many unnatural-looking or physically implausible placements. For example, a PAAK placement of a sitting motion is likely to result in either the buttocks or feet levitating above the surface of the scene due to a mismatch between the motion captured seat and the one in the scene. PACE, in contrast, alters the animation to fit the unique geometry of each scene exactly.
\section{Putting Moving Virtual Agents into Dense 3D Scenes}
In this work, we approach the following problem statement: \textit{Given a sequence of virtual agent motion and a 3D scene, create the most natural-looking and physically plausible agent-scene interaction.} Specifically, our goal is to take a given virtual agent and tailor it to a scene such that it properly fits the scene and that any interactions with the scene found in the agents motion (i.e. touching objects, sitting on a couch) match the affordances of the scene. When using cluttered or dense indoor environments, it becomes increasingly difficult to find an appropriate place to put the virtual agent. We use a data-driven method and make appropriate alterations to the virtual agents motion that result in significant improvements in the physical plausibility of agent-scene interactions and result in more natural-looking placements. 
At an intuitive level, you can imagine a motion that involves reaching for an object, but in the scene provided the object is further away than that modeled in the motion-captured environment. For the most natural-looking results, it is important that the motion is altered to reach towards the object present in the scene. Agent-scene pairings that do not alter the motion in this way will result in unnatural looking interactions that will be obvious to the user in an XR setting.

\subsection{Overview}

We present an overview of PACE in \hyperref[fig:method]{Figure 2}. A list of symbols frequently used in this work are shown in \hyperref[table:symbols]{Table 1}. Rarely used symbols are defined where they are used. In terms of notation, we define a time-series using an uppercase letter, while a lowercase letter denotes an individual frame in the time-series. 

The inputs to PACE are a set of virtual agent motions and a set of 3D scene meshes. PACE outputs a location and orientation for the virtual agent in the scene, as well as the altered pose and translation of each frame in the agents motion sequence. Each virtual agent motion sequence input, $V_b$, consists of a time series of human meshes, $v_b$, which are defined by the SMPL-X \cite{pavlakosExpressiveBodyCapture2019} parametric human body model. SMPL-X uses a human skeleton of 24 joints, all of which we can alter to tailor the motion to the scene. Each 3D scene input is represented as a static triangle mesh with vertex class labels denoting what type of object a given surface is. These class labels are helpful but not required. For the altered pose and translation outputs, PACE provides the 24 joint pose and xyz translation of each frame in the motion sequence. This can be redefined by SMPL-X to create a new motion sequence similar to the original input but properly fitting the environment.

We first take these meshes and estimate likely human scene interactions given the geometry of the mesh's surface. We then extract geometrically and semantically important frames and utilize active learning techniques to extract a diversity score for each frame. The combination of this information and the mesh geometry are combined to create the frame weights, $k$. Using the frame weights, we can find promising initial placements for the virtual agent in the scene before we optimize both the placement in the scene and the geometry of the agents motion. To optimize the motion of the virtual agent to best fit the scene, (as defined by the physical plausibility metrics we define below) we use the geometry and semantics of the scene and try to match it to the virtual agents motion, altering the specific poses in the agents motion as needed to create a match. This results in an agent-scene pairing that is as natural-looking as possible, with the motion closely matching the geometry of the scene and the interactions it contains accurately modeled.

\begin{table}[t] \label{table:symbols}
    \centering
    \begin{tabular}{| p{0.17\columnwidth} | p{0.73\columnwidth} | } 
        \hline
        \textbf{Symbols} & \textbf{Definitions}\\  
        \hline
        $V_b$ & A virtual agent motion, consisting of a time series of 3D human meshes \\ 
        \hline
        $v_b, v_p, v_\tau$ & A single 3D human mesh, an individual frame from $V_b$, the pose of that mesh, and the translation of that mesh relative to the first mesh \\ 
        \hline
        $f_c$, $f_s$ & Contact labels and Semantic labels, attributed to each vertex in a mesh, $v_b$  \\ 
        \hline
        $k$ & frame weights \\ 
        \hline
        $E_p$, $E_{alt}$ & The objective functions utilized when optimizing the motions placement and geometry \\
        \hline
    \end{tabular}
    \vspace{5.0px}
    \caption{List of symbols used and their definitions.}
    \vspace{-20.0px}
\end{table}

We choose to use an optimization method for altering the agents motion with an end-to-end deep learning model because it will \textit{always} find the ideal placement and motion geometry for the scene. In contrast to this, a deep learning model would have to learn to generalize over all possible motions and scenes, not likely finding ideal placements or motion geometry for any of them. Additionally, data does not currently exist of humans moving and interacting with dense or cluttered 3D scenes. Our optimization method also allows for flexibility to take into account other interaction metrics. For example, if not penetrating the scene is a hard constraint, users can easily alter the weight of the penetration loss and the optimization can adapt accordingly.

\subsection{Human-Scene Interaction Estimation}
To place a virtual agent motion into a scene such that any interactions present in the motion are preserved, we must first determine those interactions. Following the lead of PAAK \cite{mullenjrPlacingHumanAnimations2022}, we directly implement the POSA pretrained \cite{hassanPopulating3DScenes2021} conditional variational autoencoder (cVAE) \cite{sohnLearningStructuredOutput2015} and feed each frame of the virtual agent motion into it individually. The POSA cVAE, $f$, generates an egocentric feature map for each vertex in each mesh, $v_b$, in the virtual agent motion, $V_b$. This feature map consists of a contact label, $f_c$, and a set of semantic labels, $f_s$, which intuitively denotes a) whether that vertex should be in contact with the scene, and b), what it should ideally be in contact with, respectively. For example, if the mesh is of a person grasping an object, the vertices on the hand should have a high contact probability and highlight the object semantic label. In contrast, vertices on the other hand or on the agents back should have a very low contact probability. We represent POSA as the function $f$ in \hyperref[eq:POSA]{Equation 1} below.

\begin{equation}\label{eq:POSA}
    f : (v_b) \rightarrow [f_c,f_s]
\end{equation}

PACE  enables the use of any length of virtual agent motion in contrast to PAAK which only tested with motions 60 frames (or 2 seconds) in length. We specifically tested conducted our quantitative tests with motions from 4-15 seconds in length with 60-450 total frames. Using longer motions is only limited by the scene size, with animations longer than 15 seconds rarely having a valid fit location in the single-room, small scenes we utilized when testing PACE. We feed all of the meshes of a given virtual agent motion individually into $f$ to extract the full set of contact labels, $F_c$, and semantic labels, $F_s$, for use as inputs to our optimization methods and frame weight extractions.

\begin{figure*}[ht]
	\begin{center}
		\includegraphics[width=0.9\linewidth]{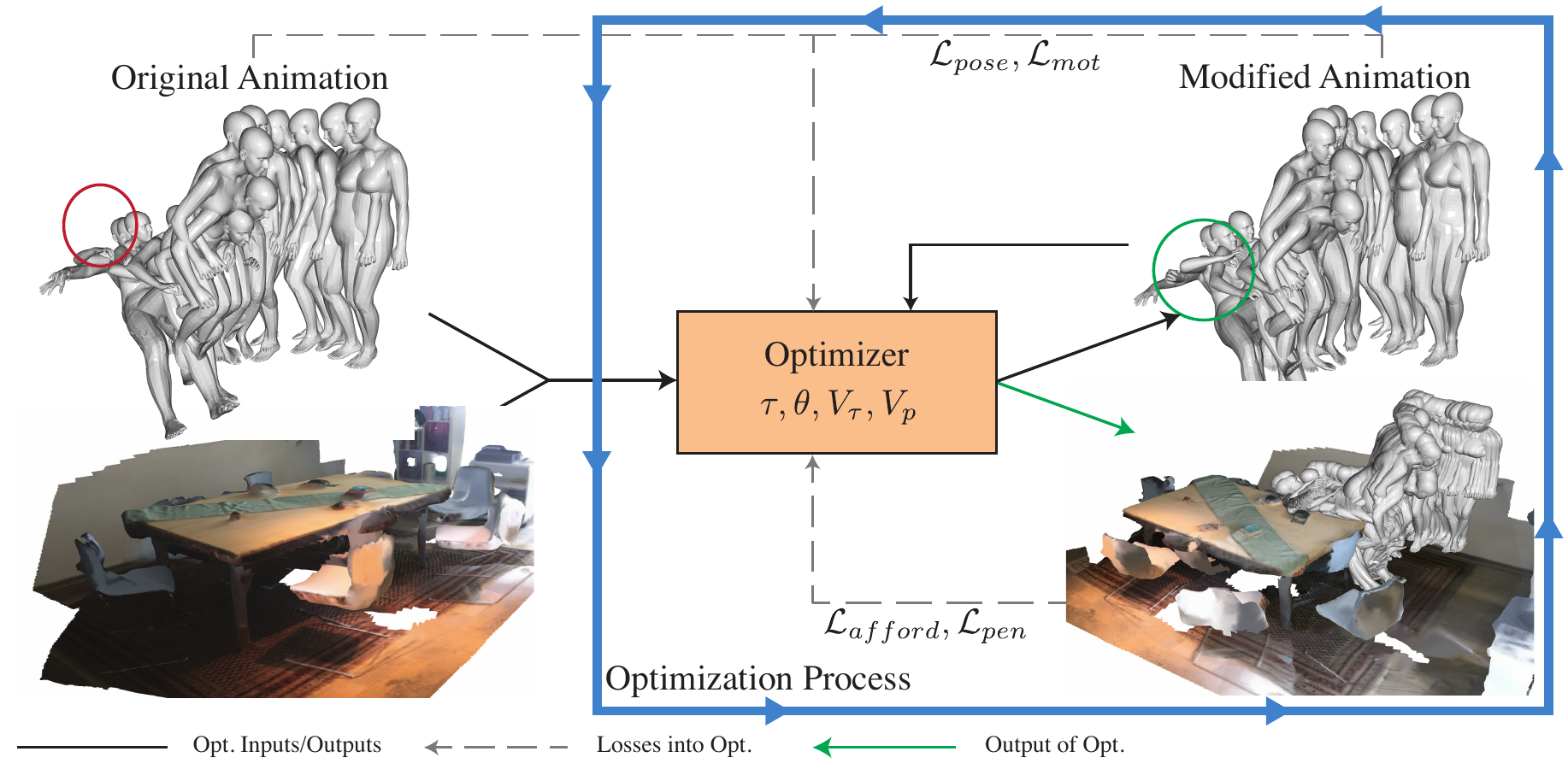}
		\caption{An overview of our novel optimization process. We iteratively modify the motion of the virtual agent while trying to find the optimal location for it in the scene. This results in a motion that better fits the scene it is being placed in, for more natural-looking and physically plausible virtual agent motion. In this example, note the green circle highlighting changes to the arm geometry so it does not collide with the table in the final placement. In the initial motion, the arm penetrated the table when the person sat down.}
		\label{fig:opt}
	\end{center}
\end{figure*}

\subsection{Frame Weighting}
We implement PAAK to extract the a frame weights for each frame in the virtual agent motion. Intuitively, finding these frame weights allows our optimization methods to focus on the most important pieces of the motion when placing it into the scene. For example, you can imagine a virtual agent which stands still for a long period before sitting down in a chair. The sitting motion would only be a few frames of the overall motion and it could be easy for the optimization process to miss these frames and instead focus on optimizing the long standing still period. However, to create the most natural-looking placement, it is extremely important that when the sitting motion occurs, it ends in a chair, while the standing period of the motion can happen almost anywhere. PAAK strives to find the frames that are most important for a natural-looking placement in the scene, $K$. PAAK consists mainly of a deep model, $g$, trained to estimate a geometric frame weighting equation. PAAK then utilizes active learning techniques to give each frame in the motion a diversity score. The combination of the model output approximating geometric and semantic cues, and the diversity score, are used to create a final frame weighting for the virtual agent motion with each frame in the motion given a weight. \hyperref[eq:al]{Equation 2} and \hyperref[eq:model]{Equation 3} below show the PAAK active frame weighting framework.

\begin{equation}
\label{eq:al}
    K = \lambda_g * \hat{K_g} + \lambda_b * W_d
\end{equation}
\begin{equation}\label{eq:model}
    g : (V_b, F_c, F_s) \rightarrow \hat{K_g}
\end{equation} 

Specifically, the PAAK deep model, $g$, maps from the virtual agent motion geometry, contact labels, and semantic labels to an estimation of a geometric frame weight definition that emphasizes motion and unique semantics. To get the final frame weights, the output of the model alongside the diversity score obtained from the models gradients are summed.

For datasets with frame rates higher than 30 fps, such as the AMASS \cite{AMASS:ICCV:2019} CMU MoCAP \cite{cmuWEB} subset with a 120 fps frame rate, we modified the PAAK algorithm to sample down to an average frame rate of 30 fps. Specifically, we utilize their deep model and extract the top 25\% of the frames by frame weight and discarded the remaining frames. To make sure we did not create large gaps in motion, we created a frame rate floor of 10 fps and retained additional frames to make sure we did not drop below this amount. Sampling down to 30 fps on average allowed our optimizer to work quicker as the number of meshes it has to calculate and the number of parameters it has to optimize are greatly decreased.

\subsection{Scene Placement and Motion Optimization}

As our main novel algorithmic contribution, we place the agent into the scene such that it makes sense in context and complete any alterations to the motion needed to better fit the scene. For the placement itself, we mainly key in on the available semantic, contact, and geometric information of the virtual agent motion as weighted by the frame weights. However, we make any alterations to the motion concurrently to finding the placement as the alterations to the motion may make a given placement significantly better. Specifically, we optimize two objective functions, $E_p$ and $E_{alt}$ where $E_p$ finds the optimal location, $\tau$, and rotation, $\theta$, in the scene for the virtual agent and where $E_{alt}$ finds the optimum alterations to the agents motion to maximize the viability of the placement without generating too much movement as to hurt the perceived realism of the motion. As in \cite{mullenjrPlacingHumanAnimations2022} and \cite{hassanPopulating3DScenes2021}, $E_p$ minimizes the sum of an affordance loss, $\mathcal{L}_{afford}$ and a penetration loss, $\mathcal{L}_{pen}$, calculated for each frame weighted by the frame weights, $k_a$.

\begin{equation}
    E_p(\tau, \theta) = \sum^{|k|}_{i=0}{k_{i}*[\mathcal{L}_{afford, i}} + \mathcal{L}_{pen, i}]
\end{equation}

Intuitively, $\mathcal{L}_{afford}$ is minimized when the distance to the scene is small for vertices with a high probability of contact using $f_c$ and when the semantic label $f_s$ matches the semantics of the object vertices are in contact with. $\mathcal{L}_{pen}$ heavily penalizes placements that result in the agents motion penetrating the scene as that is a key factor in physical plausibility. The frame weights, $k$, allows the optimizer to accurately find a placement that focuses on the key interactions in the virtual agent motion.

For the alteration objective, $E_{alt}$, two more loss functions are added with the intention of maintaining consistency through the virtual agent motion. $E_{alt}$ minimizes the sum of a pose loss, $\mathcal{L}_{pose}$, and a motion loss $\mathcal{L}_{mot}$. $\mathcal{L}_{pose}$ is minimized when the pose of each individual mesh in the motion is close to the original pose, thus minimizing unrealistic poses. $\mathcal{L}_{mot}$ is minimized when the motion between two frames is the same as from the original virtual agent motion, thus minimizing the probability of overly large single motions between frames. The weighting for these two loss terms are set such that an individually large difference in pose with a number of small differences in motion is less costly than large changes in motion. This creates an effectively connected nature where the optimizer favors a smooth motion towards the goal pose over a sudden change in pose. The optimizer for $E_{alt}$ operates across the translation of each mesh in the virtual agent motion and the pose of each skeleton in each frame of the virtual agent motion. \hyperref[eq:Lpose]{Equation 6} shows the calculation pose loss, taking effectively the squared error of the pose for each frame. Similarly, \hyperref[eq:Lmot]{Equation 7} shows the calculation of the motion loss, which is the squared error of the difference between each sequential pose and translation in the motion sequence. We added a discount factor, $\lambda_{\tau}$ for the translation component of the loss as we found we achieved better results when emphasizing it less, but worse results if we removed it entirely. This process can be seen visually in \hyperref[fig:opt]{Figure 3.}

\begin{equation}\label{eq:Ealt}
    E_{alt}(V_\tau, V_p) = \sum^{|k|}_{i=0}{k_{i}*[\mathcal{L}_{pose, i}} + \lambda_{mot} * \mathcal{L}_{mot, i}]
\end{equation}
\begin{equation}\label{eq:Lpose}
    \mathcal{L}_{pose} = \sum^{|k|}_{i=0}{(v_{p,i} - \hat{v_{p,i}})^2}
\end{equation}
\begin{multline}\label{eq:Lmot}
        \mathcal{L}_{mot} = \sum^{|k|}_{i=0}{(diff(v_p)_i - diff(\hat{v_p})_i)^2} + \\ \lambda_{\tau}[\sum^{|k|}_{i=0}{(diff(v_\tau)_i - diff(\hat{v_\tau})_i)^2}]
\end{multline}

\begin{figure*}[ht]
	\begin{center}
		\includegraphics[width=\linewidth]{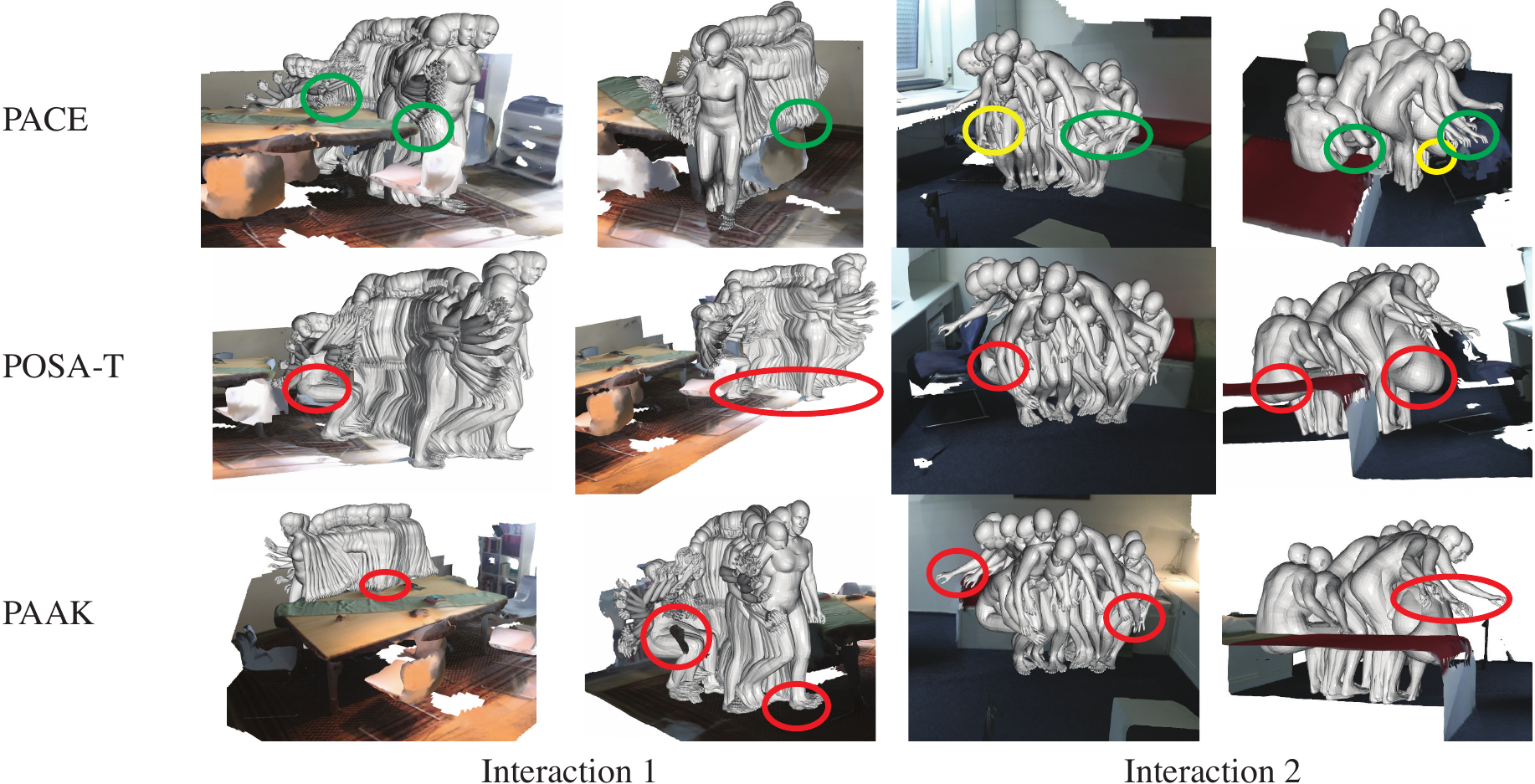}
		\caption{Comparisons on placing the same virtual agent into the same scene across PACE, POSA-T, and PAAK. Note that two angles of each placement are provided. For the first placement, PACE is the only one that maximizes the interaction with the environment by tailoring the agents motion to maneuver it around the chair to the right of the table, placing the hand on it as a guide as a real person might. For POSA-T, the placement not only puts the agent perpendicular to the chair, but then has it wonder off the scene where its interactions make less sense. PAAK is in a middle state where it maintains contact with the environment but penetrates the chair and table and floats above the ground surface. For placement 2, PACE shows the most probable interaction given the virtual agent motion provided. It interacts with the chair and the bed, bracing its movement to the chair with the hand. It does however result in an awkward yet still valid seating position. For POSA-T, the second seating position is completely ignored, which makes sense due to its lack of frame weighting. PAAK finds a valid placement for the virtual agent but the hand placements are awkward and do not contact the scene in the way a real human might.}
		\label{fig:qual}
	\end{center}
\end{figure*}

$E_{alt}$ when used in combination with $E_p$ is especially powerful as the optimizer can fit the virtual agent motion to the scene while retaining the essence and natural look of the real motion sequence. For example, in a scenario where the virtual agent is sitting in a chair that is smaller than that which was recorded in the original motion capture, the feet of the agent would normally go through the floor. However, when using $E_{alt}$ and $E_p$ together, the actual poses of the agent can change to make sure the feet do not penetrate the floor, while maintaining a smooth and natural-looking sitting motion. Without $E_{alt}$ there are many scenarios where the virtual agent motions do not closely fit the needs of the scene resulting in penetrations, floating feet, or inaccurate affordances. These errors result in physically implausible and unnatural-looking placements for the virtual agents that quickly break users immersion and satisfaction.

We specifically use a Pytorch implementation of L-BFGS with Strong Wolfe line search as our optimizer following \cite{hassanPopulating3DScenes2021} and \cite{mullenjrPlacingHumanAnimations2022}. A learning rate of 1 is used and the optimization executes for 10 steps.

\subsection{Physical Plausibility Metrics}

Quantitatively measuring the physical plausibilty of interactions generated by PACE is important for validating its results. For this, we use the non-collision and contact metrics originally defined by Zhang et al. \cite{zhangGenerating3DPeople2020} and used by Hassan et al.\cite{hassanPopulating3DScenes2021}, and Mullen et al. \cite{mullenjrPlacingHumanAnimations2022}. Each metric is defined on the human mesh level. Given the body mesh, $v_b$, the scene mesh, $m_s$, and a scene signed distance field (SDF) that stores distances for each voxel, the non-collision score is computed as the ratio of body mesh vertices with positive SDF values divided by the total number of vertices. In contrast, the contact score is calculated as 1 if at least one vertex of $v_b$ has a non-positive distance value to the scene.

\section{Experiments}
\subsection{Datasets and Baseline}

For our PROX based evaluations, we randomly sample 10,000 motions from the 8 training scenes of PROX to use as our virtual agents. We then place these virtual agents into the 4 testing scenes. For our perceptive user study, these placement results are then shown to human raters in comparison with a baseline or ablative method. The rates are shown two placements in the same scene and asked to pick the more realistic of the two videos. For our quantitative evaluations of the physical plausibility metric all of the methods are used to generate results to be compared directly.

For our AMASS based evaluations, we randomly sample relevant virtual agent motion sequences from the CMU MoCap dataset. This means that we filter out motions that involve dancing, climbing, or those that would otherwise be unnatural in a cluttered indoor environment. We then use a HoloLens augmented reality headset to sample a local scene and populate that scene with a virtual agent using PACE. This was done as a proof of concept for how this could work in use. Additionally, we will be releasing a dataset we created using the AMASS/CMU dataset and scenes from PROX and Matterport3D \cite{Matterport3D}.

\begin{figure*}[ht]
	\begin{center}
		\includegraphics[width=\linewidth]{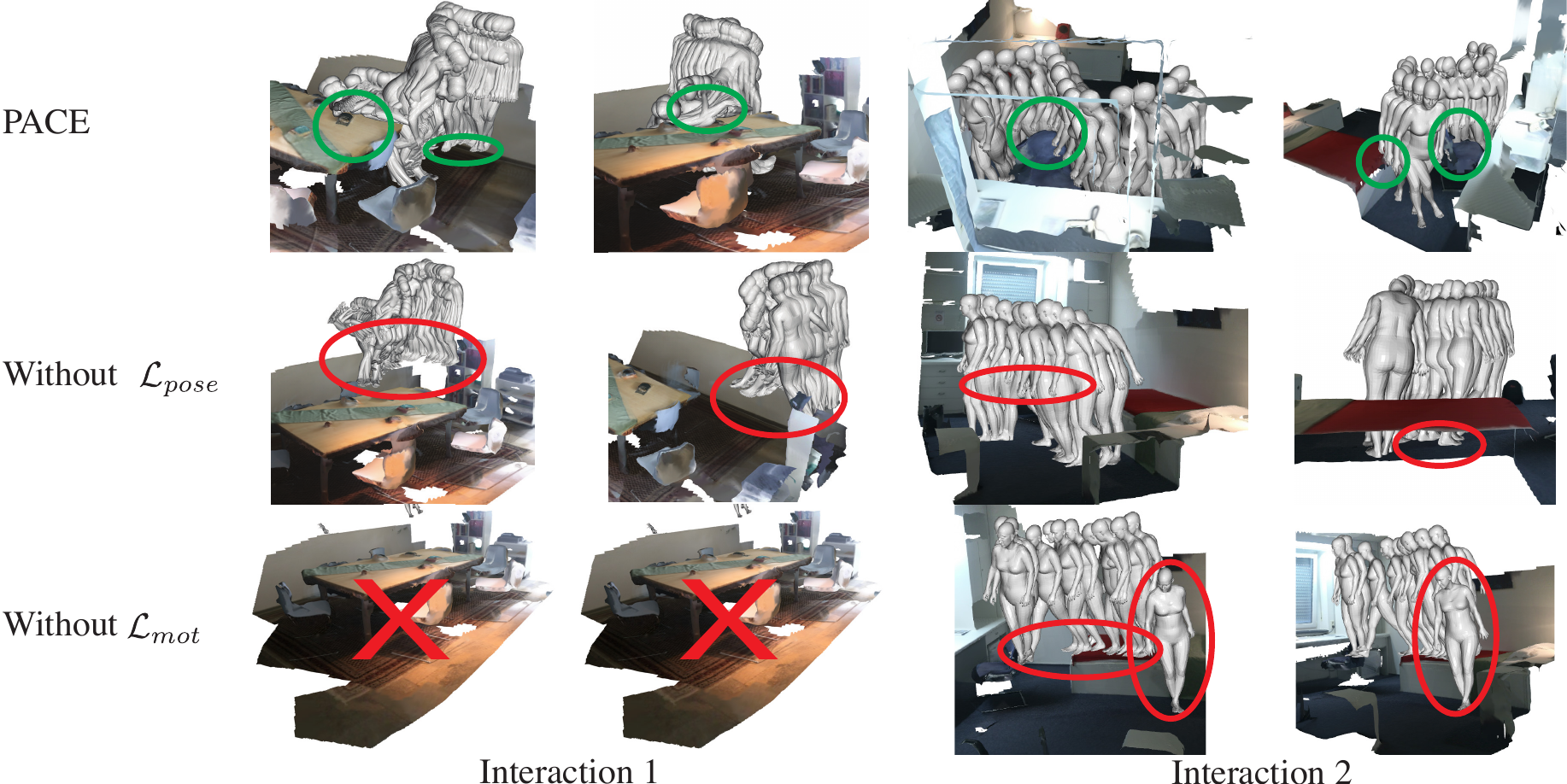}
		\caption{Comparisons on placing the same virtual agent into the same scene across PACE, and two ablation studies. Specifically, we attempt to remove the $\mathcal{L}_{pose}$ and $\mathcal{L}_{mot}$ terms and evaluate the results. Note that two view angles of each interactions are provided. For the first interaction, notice how PACE places the agent such that it has direct, non-penetrative contact with the table and the ground that matches the motion of the agent. Conversely, removing $\mathcal{L}_{pose}$resulted in the agent seeming to float out of the scene, and contort in awkward ways, i.e. a non-plausible interaction. Removing $\mathcal{L}_{mot}$ resulted in no visible placement as each individual mesh was separated and thrown far from the scene. For placement 2, PACE successfully navigates the virtual agent around the chair, and even results in contact between the hand and the chair for support. In contrast, the virtual agents motion without $\mathcal{L}_{pose}$ resulted in some awkwardly leaned poses and a lack of contact for some of the meshes. Moreover, the motion sequence without $\mathcal{L}_{mot}$ became separated into pieces, not resembling a valid motion at all. Overall, $\mathcal{L}_{pose}$ and $\mathcal{L}_{mot}$ improve the plausibility of the interactions.}
		\label{fig:qual-ablate}
	\end{center}
\end{figure*}

We compare PACE with POSA, PAAK, a motion synthesis method, and ablations of PACE. We describe the baselines in greater detail below:

\textbf{POSA-T.} Hassan et al. \cite{hassanPopulating3DScenes2021} propose a method that places a single human mesh into a scene given its affordances. As this approach does not consider motion, we use PAAK's \cite{mullenjrPlacingHumanAnimations2022} POSA-T baseline which is modified to sum the placement loss over all the meshes in the motion sequence. It is labeled POSA-T for the addition of the time dimension.

\textbf{PAAK.} For PAAK, Mullen et al. \cite{mullenjrPlacingHumanAnimations2022} selects ``keyframes" from a virtual agent to focus on when optimizing the placement of the agent into a scene. 

\textbf{Motion Synthesis.} Wang et al. \cite{wangSynthesizingLongTerm3D2021} propose a motion synthesis method that accounts for the affordances of the scene in its motion creation. We used their approach as designed. We call this baseline Motion Synthesis.

\subsection{Evaluation}
A qualitative evaluation of PACE alongside PAAK and POSA-T can be found in \hyperref[fig:qual]{Figure 4}. Generally, we found that PACE generated improved interactions and fittings over the baselines, especially with regard to the details of each interaction. POSA-T was especially prone to invalid placements or interactions, like having the agent sitting in midair. PAAK placements tended to decrease in quality as the length of the virtual agents motion increases while its interactions were clumsy and inexact. This is likely due to its inability to alter the motion itself. This prevented it from tailoring the motion to the small, cluttered environments in the PROX dataset. These phenomenon can be seen in \hyperref[fig:qual]{Figure 4}. Anecdotally, as scene complexity increased, valid fittings tended to decrease. As a simple example, if you imagine a scene with a chair placed such that its right side is touching a wall, a motion that goes from sitting to standing, before immediately turning right and walking, will not have a valid fit location. These types of issues become more common as scene complexity increase, and are especially relevant for PAAK which cannot make small changes to the motion to accommodate the scene.

\begin{table}[t] \label{table:metric}
    \centering
    \begin{tabular}{l c c } 
        \hline\hline
        & Non-Collision $\uparrow$ & Contact $\uparrow$ \\ [0.5ex] 
        \hline\hline
        POSA-T & 0.983 & 0.733 \\ 
        \hline
        PAAK & 0.986 & 0.784 \\
        \hline
        \textbf{PACE} & \textbf{0.996} & \textbf{0.929} \\
        \hline
        PACE w/o $\mathcal{L}_{pose}$ & 0.995 & 0.610\\
        \hline
        PACE w/o $\mathcal{L}_{mot}$ & 1.00 & 0.340\\
        \hline
    \end{tabular}
    \vspace{5.0px}
    \caption{Evaluation of the physical plausibility metrics. A higher score is better for both the non-collision score and the contact score. PACE outperforms both of our baselines, especially in the contact metric. Moreover, PAAK results in a higher number of collisions between the virtual agent and the obstacles, while also having less meaningful interactions with the scene. Additionally, while the ablative methods without some of our loss terms perform well in the non-collision metric, they resulted in far lower contact scores than PACE.}
\end{table}

\textbf{Physical Plausibility.} Following the procedures utilized by \cite{hassanPopulating3DScenes2021, zhangPLACEProximityLearning2020, zhangGenerating3DPeople2020, mullenjrPlacingHumanAnimations2022}, we take 100 virtual agents and place them into the 4 test scenes of PROX. With the placements providing a body mesh relative to the scene, we use that and a scene signed distance field (SDF) to compute a non-collision score and contact score as defined originally in \cite{zhangGenerating3DPeople2020} and explained in section 3. The results are shown in \hyperref[table:metric]{Table 2}. A high non-collision score denotes that the meshes in each motion sequence do not penetrate the scene. PACE is an improvement over the baselines in the non-collision score.

The contact score is calculated as 1 if at least one vertex of the mesh is in direct contact with the scene. PACE is a significant improvement over the baselines, showcasing a unique ability to bend the virtual agents motion towards the geometry of the scene. The decrease in performance with PAAK and POSA-T are likely due to mismatches between the motion geometry in the scene geometry in things like seat size or exact object distances. For example, when standing up after sitting, if the seat size was not correct, the feet could be positioned above the floor instead of in contact with it. In contrast, PACE is able to ensure contact at each frame in the motion sequence, maximizing the contact metric. These results show that PACE creates more physically plausible placements than the baselines.

\begin{figure}[t]
	\begin{center}
		\includegraphics[width=0.8\columnwidth]{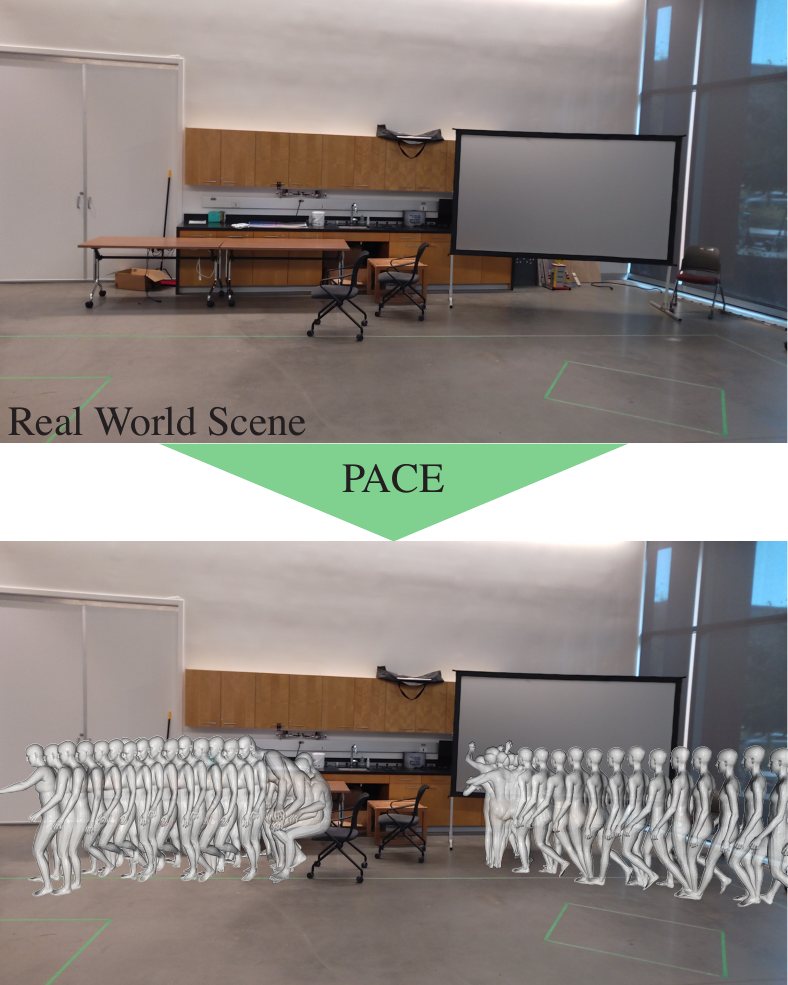}
		\caption{PACE evaluated in a real-world scene using HoloLens. Notice how the virtual agent smoothly navigates around the obstacles in the real-world environment.}
		\label{fig:hololens}
	\end{center}
\end{figure}

\textbf{Ablation Results.} We ablated PACE against the use of the $\mathcal{L}_{pose}$ and $\mathcal{L}_{mot}$ loss functions and compared the results qualitatively and with the physical plausibility metrics. For the physical plausibility metrics, both of these ablations resulted in similar or less collisions than PACE. However, the contact metric rapidly decreases. This makes sense when visualizing the results qualitatively in \hyperref[fig:qual-ablate]{Figure 5}. When ablating against $\mathcal{L}_{pose}$, we notice that the poses of the human itself rapidly decomposes to look unnatural and disfigured. The virtual agent itself also moves rapidly away from the scene as if taking flight. This allowed the optimizer to negate any penetration with the environment, at the cost of contacting it in any way and retaining realism in the motion. The results are also poor when ablating against $\mathcal{L}_{mot}$. As the translational location and motion of each mesh relative to the previous frame are no longer enforced, the virtual agent would often be moved entirely out of the frame by the optimizer. Some individual meshes would sometimes remain when there was important semantic contact.

\textbf{HoloLens Study.} We developed an augmented reality system for the Microsoft HoloLens where the application samples the geometry of its current environment, and places a virtual agent into it using PACE. \hyperref[fig:hololens]{Figure 6} shows an example of this implementation. We found a number of unique challenges, mostly related to the inaccuracies in the HoloLens room scans. It would box some things off incorrectly, resulting in conflicts. The HoloLens also does not classify the objects in the room semantically, making it more difficult to place virtual agents that want to specifically key in on this information. For example, a virtual agent with a laying interaction may lay on the floor despite there being a bed nearby. Anecdotal feedback suggests that users were impressed by the natural-looking motion of the virtual agents, but were not completely convinced they could be real people. Specifically, users noted that virtual agents seemed to often be walking around aimlessly which a real human would not typically do. After precomputing the scene and motion sequences, the final virtual agents can be populated into the HoloLens at 30 frames per second.

\begin{figure}[t]
	\begin{center}
		\includegraphics[width=0.8\columnwidth]{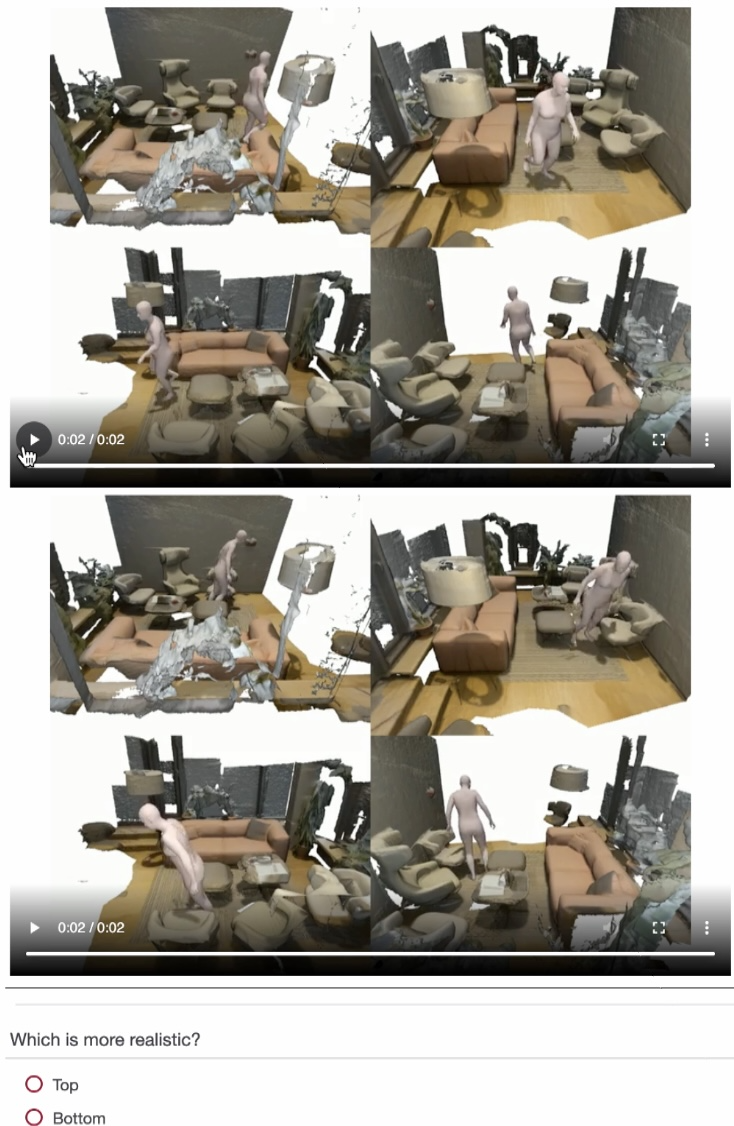}
		\caption{Example question shown to participants in our user study. Two videos are presented, one corresponding to PACE, the other corresponding to a baseline. Users would play each video as many times as they wished and would see the virtual agent move through the scene. Users would then have to choose which video was more realistic or plausible.}
		\label{fig:user-study}
	\end{center}
\end{figure}

\subsection{User Study}
We conduct a web-based user study to determine the quality of the virtual agent-scene pairings PACE creates compared to those created by the baselines.

\textbf{Procedure.}
We follow the protocols of Hassan et al. \cite{hassanPopulating3DScenes2021}, Zhang et al. \cite{zhangPLACEProximityLearning2020}, and Mullen et al. \cite{mullenjrPlacingHumanAnimations2022}. The study consisted of 60 questions taking participants approximately 30 minutes to complete. For each question, participants were shown two clips of a virtual agent placed into a cluttered 3D scene. They would play these two clips and decide which one they perceived to look more natural or realistic. An example user study question can be seen in \hyperref[fig:user-study]{Figure 7}. 

To generate our clips, we used the 4 testing scenes from the PROX dataset. We then took 100 virtual agent samples from the PROX training set for placement into the scenes. All of the methods, with the exclusion of the synthesis method, begin with a grid of potential placement locations and orientations in the scene with each location testing rotations every 30 degrees. We then find the optimal placement and alteration by minimizing $E_p$ and $E_{alt}$. We then render each agent-scene pair into a video from four different angles so the human raters can get a sense of the relationships between the agent, its motion, and the scene. 

We had a total of 400 agent-scene pairings for each method, 100 for each of the 4 utilized scenes. Users were shown a selection of these as to not make the study prohibitively long. Specifically, 5 clips from each of the four scenes, for each method, were selected. As we compared against 3 baseline methods, this resulted in 60 total questions per participant. Participants were not told which clip was from which method. Moreover, we made sure that PACE was not always placed above or below the baseline to prevent any selection bias. As the selection of clips was random, some clips were viewed and rated more frequently than others by participants.

\begin{table}[t] \label{table:direct}
    \centering
    \begin{tabular}{l c c } 
        \hline\hline
        & Baseline $\downarrow$ & PACE $\uparrow$ \\ [0.5ex] 
        \hline\hline
        POSA-T & 32.1\% & \textbf{67.9\%} \\ 
        \hline
        Motion Synthesis & 18.8\% & \textbf{81.2\%} \\
        \hline
        PAAK & 40.7\% & \textbf{59.3\%} \\
        \hline
    \end{tabular}
    \vspace{5.0px}
    \caption{PACE compared to POSA-T, Motion Synthesis \cite{wangSynthesizingLongTerm3D2021}, and PAAK. Subjects are shown pairs of motions placed into a 3D scene by PACE and by a baseline and must choose the most realistic one. A higher percentage indicates the scene-agent pairing subjects deemed more plausible and exhibits better interaction. Users preferred PACE over all of the baseline methods.}
\end{table}

\textbf{Evaluation.}
42 participants took part in our study. They were recruited by web advertisements and word of mouth. We did not record demographic information as to avoid any identifying information being collected. Additionally, unlike similar studies looking at emotion or gestures, demographic background should not impact whether a virtual agent's motion looks physically plausible.
As each participant responded to 60 questions, we had a total of 2520 responses. The results are shown in Table 2. 

PACE is an improvement over PAAK, with participants preferring it 59.3\% of the time. Participants sometimes found it difficult to fully notice the difference in some placements due to the low resolution of the video. Where users did see the difference is in the even more cluttered or smaller rooms where PAAK would often generate interactions that would penetrate the scene (i.e. collisions), or be entirely at an elevation that does not make realistic sense, like walking on a bed. This makes sense as PAAK was unable to modify the motion of the agent to properly respond to the unique characteristics of the environment, effectively shoehorning in motion that does not accurately fit into the scene. Participants perceived PACE as more natural-looking than POSA-T 67.9\% of the time. This is likely because POSA-T created much worse interactions with common occurrences including the virtual agent levitating or sitting in mid-air. The most plausible explanation for this being POSA-T's lack of frame weights and inability to alter the motion of the agent to properly fit the scene. Users strongly disliked the virtual agents created by the motion synthesis method, preferring PACE 81.2\% of the time. The most likely explanation for this is the motion synthesis method not consistently generating smooth, plausible motion. Individual frames of the motion generated by the synthesis method looked great but when it played as a video, like that shown to participants, the motion often looked unrealistic. Specifically, legs frequently did not move in a natural way when walking. This result emphasizes the importance of using motion-captured virtual agents to create the most natural-looking agent-scene pairings.

\section{New Virtual Agent Movement Dataset}
Using the AMASS \cite{AMASS:ICCV:2019} CMU MoCap \cite{cmuWEB} dataset, we used PACE to create a new dataset of virtual agent placements in cluttered indoor 3D scenes. We are releasing this dataset of tens of thousands of agent-scene pairs. We use the fitting loss, $E$, as a quality marker on the fitting and save pairings with a loss below a threshold value. Additionally, we filter the placements by the physical plausibility metric, ensuring that pairings with poor plausibility are not included. This filtering process allows us to filter out virtual agents that are poor fits for a given scene and would need an unnatural amount of alteration to fit properly. Completing this process with the full AMASS/CMU MoCap dataset alongside the PROX and Matterport3D-R \cite{zhangPLACEProximityLearning2020} room scans results in tens of thousands of samples. These samples are labeled by the original CMU MoCap data labels so users can filter by motion type. The motions included in our dataset include walking, jumping, stretching, sitting, navigating, among others. This dataset would be released at the time of publication.
\section{Conclusions, Limitations, and Future Work}

In this paper, we propose PACE, a novel method for generating placements for virtual agents into a dense cluttered 3D scene with accurately modeled human-scene interactions and tailored motion. Our key insight is that tailoring the motion of the virtual agents is essential to creating agent-scene pairings that are natural-looking, especially in dense or cluttered environments. Human raters preferred PACE agent placements over existing methods including PAAK and POSA. Additionally, PACE agent-scene pairs are quantitatively rated as more physically plausible than existing methods.

\textbf{Limitations and Future Work.} Note that PACE does not \textit{always} create natural placements. There are still cases where a given motion is simply too long or large to fit in a given environment. Additionally, if there is no semantic match the placement can look unnatural. For example, a sitting motion will not look right in a scene without any chairs. PACE is currently limited to static 3D scenes, although future work can explore the use of dynamic scenes. Our optimization method can sometimes miss good placements as it relies on a grid of initial placements before optimizing the best ones. For example, initial placements with heavily penalized penetrations could become the best available with further optimization to the motion sequence. We limited this due to processing time but more compute could enable improved agent-scene pairings from PACE. This leads us to the largest limitation of PACE, the amount of processing time it requires. Future work can look into placement proposal to limit the number of initial placements PACE must attempt to optimize. This would work similar to the way a region proposal network works for object detection models. An additional interesting future direction would extend PACE to model human-human interactions when placing multiple animations into a scene. More testing with real-world AR and VR headsets is needed to determine how users respond to the virtual agents added by PACE.

\textbf{Acknowledgements.} This material is based upon work supported by the National Science Foundation Graduate Research Fellowship Program under Grant No. DGE 1840340. It is also supported by ARO Grant W911NF2110026  and U.S. Army Cooperative Agreement W911NF2120076. Any opinions, findings, and conclusions or recommendations expressed in this material are those of the author(s) and do not necessarily reflect the views of these funding agencies.


\bibliographystyle{abbrv-doi}

\bibliography{main}
\end{document}